Artificial Intelligence and Machine Learning in the Development of Vaccines and Immunotherapeutics

Yesterday, Today, and Tomorrow

Elhoucine Elfatimi[1], Yassir Lekbach[1]; Swayam Prakash[1]; Lbachir BenMohamed[1, 2, 3*]

[1]Laboratory of Cellular and Molecular Immunology, The Gavin Herbert Eye Institute, University of California, Irvine, College of Medicine, Irvine, CA 92697-4375; [2]Institute for Immunology & [3]Chao Family Comprehensive Cancer Center, University of California, Irvine Medical Center, CA 92868, USA.

Running Title: Artificial Intelligence in Vaccines and Immunotherapeutics Development

*Address correspondence and reprint requests to Dr. Lbachir BenMohamed. Laboratory of Cellular and Molecular Immunology, The Gavin Herbert Eye Institute, University of California, Irvine, College of Medicine, Hewitt Hall, 843 Health Sciences Road, building 843, 2nd Floor, Room 2032, Irvine, CA 92697, ZOT 4390. Phone: 949-824- 8937. Fax: 949-824-9626. E-mail: Lbenmoha@uci.edu.

Word counts: Abstract (300), Text (11204)

# Abstract

The development of vaccines and immunotherapies against infectious diseases and cancers has been one of the major achievements of medical science in the last century. Subunit vaccines offer key advantages over whole-inactivated or attenuated-pathogen-based vaccines, as they elicit more specific B- and T-cell responses with improved safety. However, developing subunit vaccines is often cost and time-consuming and may not predict fast, strong, and long-lasting immunity, limiting their ability to rapidly counter apparent growing emerging pandemics and cancers. In the past, the development of vaccines and immunotherapeutics relied heavily on trial-and-error experimentation and extensive *in vivo* testing, often requiring years of pre-clinical and clinical trials. Today, artificial intelligence (AI) and deep learning (DL) are actively transforming vaccine and immunotherapeutic design, by (*i*) offering predictive frameworks that support rapid, data-driven decision-making; (*ii*) increasingly being implemented as time- and resource-efficient strategies that integrate computational models; systems vaccinology and multi-omics data to better phenotype, differentiate, and classify patients diseases and cancers; predict patients' immune responses and identify the factors contributing to optimal vaccine and immunotherapeutic protective efficacy; (*iii*) refining the selection of B- and T-cell antigen/epitope targets to enhance efficacy and durability of immune protection; and (*iv*) enabling a deeper understanding of immune regulation, immune evasion, immune checkpoints, and regulatory pathways. The future of AI and DL points toward (*i*) replacing animal preclinical testing of drugs, vaccines, and immunotherapeutics with computational-based models, as recently proposed by the United States FDA; and *(ii)* enabling real-time in vivo modeling for immunobridging and prediction of protection in clinical trials. This may result in a fast and transformative shift for the development of personal vaccines and immunotherapeutics against infectious pathogens and cancers.

## Introduction

The relationship between artificial intelligence (AI) and immunology is intricate and transformative (1). AI models are increasingly used to enhance our understanding of the cellular and molecular components of the immune system, leveraging computational power to identify complex interactions across immune pathways and disease contexts (2-4). Yesterday, immunological research relied primarily on experimental trial-and-error and time-consuming laboratory assays. These traditional methods, though foundational, were limited in scope and scalability.  In the early days of computational immunology, traditional mathematical models were the primary tools for simulating immune responses. These early models relied on simplified assumptions and static parameters, often failing to account for the inherent complexity and variability of immune interactions across diverse populations. Prior to the implementation of artificial intelligence, immune system modeling was limited by computational constraints, small datasets, and the absence of real-time adaptability. This 'yesterday' phase laid the groundwork for today's AI-driven approaches, highlighting both the potential and the limitations of classical immunological simulations. AI-driven models provide insights into immunotherapeutic clinical trials, improving their delivery and efficacy; however, AI approaches can sometimes be misleading when they fail to consider the full complexity of immunological interactions, especially when focusing on isolated mechanisms without integrating broader immunological networks (5). When experimental findings challenge existing AI-driven predictions, computational scientists must refine their models, ensuring they remain aligned with empirical immunology. While AI-driven hypothesis generation can be insightful, practical implementation remains a challenge. Many AI models lack robustness in real-world applications due to oversimplifications, biases in training datasets, and an inability to capture the full scope of immune system variability (6). Moreover, certain AI models struggle to provide reliable experimental validation pathways, leading to limitations in their clinical applicability (7).

There is significant debate among immunologists regarding: **(i)** the role of AI-driven models in immunology (referred to in this report as immuno-AI) and **(ii)** how AI can be effectively employed to understand complex non-linear immunological systems (7). Immunology is inherently dynamic, involving multi-scale biological interactions that require advanced computational approaches for effective modeling. Today, AI and machine learning (ML) are at the forefront of immunological research, enabling unprecedented capabilities in immune system simulation, epitope mapping, and immunotherapy design. AI has been instrumental in studying immune responses in viral infections (e.g., HIV-1) (8), cancer (e.g., chronic myeloid leukemia) (9), and autoimmune disorders (e.g., Alzheimer's disease) (10). Additionally, AI models have recently been employed to examine how multiple viruses interact and replicate within the same host cell (e.g., HIV-1 and HSV in CD4$^+$ T cells or dendritic cells) (11). Traditional vaccine development relies on labor-intensive antigen screening, which can result in suboptimal formulations. AI-driven approaches have emerged as powerful tools to streamline epitope selection by leveraging diverse datasets, including single-cell RNA sequencing, structural protein modeling, and immune response profiling (12). Recent studies have demonstrated the potential of AI-powered frameworks, such as generative models and deep learning techniques, in accelerating vaccine design by predicting immune responses and optimizing multi-epitope formulations (13, 14).

Artificial intelligence systems process large-scale immunological datasets, translating them into predictive frameworks that provide logical insights into immune responses, facilitate vaccine development, and guide novel immunotherapy strategies (**Fig. 5**). Reflecting the growing confidence in AI-driven approaches, the U.S. FDA recently announced plans to phase out certain animal testing requirements and replace them with AI-based computational models and human organoid systems to enhance preclinical evaluation and improve translational relevance (15). AI also excels in generating counterintuitive insights, uncovering immune interactions that may not be evident through conventional in vitro or in vivo experimental assays (16). However, many immunologists remain skeptical about fully integrating AI-driven approaches, as computational models often differ from traditional wet-lab methodologies (17). The emerging field of immuno-AI aims to bridge this gap by fostering interdisciplinary collaboration between AI researchers and immunologists. AI models are playing a transformative role in vaccine and immunotherapy development by enabling more precise immune response predictions. Deep learning-based classifiers, such as convolutional neural networks (CNNs), have been utilized to differentiate protective versus non-protective immune responses (18). By integrating AI-powered biomarker discovery, these models contribute to refining vaccine formulations for enhanced and long-lasting immunity. AI-driven immunogenicity prediction frameworks, validated through experimental approaches, demonstrate their potential in optimizing vaccine efficacy and guiding next-generation immunotherapy strategies (19).

Artificial intelligence plays an essential role in deciphering the non-linear dynamics of the immune system. Most immunological interactions involve complex feedback loops and multi-scale regulatory mechanisms. The immune system is highly non-linear, meaning that small perturbations in immune signals can lead to disproportionate effects on immune responses. A key example is T-cell activation in response to antigen (Ag) concentration (20). T cells exhibit threshold-based responses, where minimal antigen exposure triggers no response, while increased antigen concentration induces an exponential increase in activation. However, at extremely high antigen levels, T-cell responses may decline due to immune exhaustion mechanisms (e.g., $PD1^+CD8^+$ or $TIM^+CD8^+$ T cells) (21). While immunologists have long studied dose-response relationships, AI enables a more detailed exploration of how immune responses dynamically shift across different conditions. The complexity of cytokine networks, chemokine interactions, and immune checkpoints makes AI-based models particularly valuable for predicting immune system behaviors in health and disease (22).

One of the greatest challenges AI researchers face in immunology is capturing the full complexity of immune phenomena. Immunologists themselves often struggle to fully decipher immunological mechanisms due to current limitations in experimental assays and data interpretation. AI models must address critical gaps in immune system modeling, including predicting the speed of immune responses, analyzing T-cell subset differentiation (Th1, Th2, Th17), and simulating immune activation under diverse conditions (19). AI has been applied to $CD4^+$ T cell differentiation modeling, but many studies remain theoretical and lack robust experimental validation (23). To overcome these challenges, AI-driven models must balance complexity with interpretability. Both immunologists and AI scientists often rely on simplified models that capture essential immune dynamics without excessive abstraction. AI enables pattern recognition, predictive analytics, and experimental design improvements, allowing researchers to identify key immune parameters for hypothesis testing (24). However, over-reliance on AI without validation through empirical immunology can lead to misleading conclusions. Base immune-like AI models should be viewed as complementary to wet-lab immunology rather than as standalone predictive tools. The rapid expansion of AI applications in immunology has been facilitated by advances in user-friendly computational tools. These platforms enable immunologists to integrate AI-driven insights into their research without requiring extensive programming expertise. AI models often start with fundamental immunological assumptions, which are refined through iterative machine learning approaches. However, AI-driven predictions must be carefully scrutinized to ensure their applicability to real-world immunological settings. While AI models have led to numerous breakthroughs, like AstraZeneca's usage of AI to inform cancer drug trials through Immune, challenges persist in translating computational insights into clinically actionable immunological strategies.

A growing number of AI approaches aim to simulate adaptive immune responses, including B-cell and T-cell activation (25, 26). AI-based immune simulations can generate realistic predictions that align with experimental findings, but many models remain overly simplified. One limitation is the failure to account for the dynamic interplay between multiple immune components (27). AI simulations should prioritize mechanistic insights that can be experimentally validated rather than focusing solely on predictive accuracy. Immunologists often express concerns about AI models omitting critical immune variables. However, just as experimental studies focus on a subset of parameters (e.g., antigen load, cytokine levels, and immune cell counts), AI-driven approaches can provide valuable insights by identifying essential immune interactions (27, 28). Simple AI models can be particularly useful for ruling out ineffective immune mechanisms rather than solely predicting successful outcomes. AI-based models are also increasingly employed in vaccine design and immunotherapy development (29-31). By optimizing clinical trial parameters, AI can reduce the number of experimental groups, refine vaccine formulations, and optimize immune checkpoint inhibitor protocols (30). AI can guide vaccine development by predicting immunogenic epitopes, assessing immune memory responses, and evaluating vaccine efficacy across diverse populations (32, 33). Tomorrow, the integration of advanced AI frameworks, including generative models, multi-modal learning, and interpretable machine learning, will further accelerate the design of personalized vaccines and immunotherapies, creating a future where immune responses can be simulated and optimized in silico before entering clinical pipelines.

In summary, AI models are transforming immunological research by facilitating experimental design, identifying key immune interactions, and optimizing therapeutic strategies (34, 35). AI enables rapid hypothesis generation and computational simulations that guide experimental immunology in new directions (36). Under a broader "systems immunology" perspective, AI helps immunologists (i) to focus investigations for rapid discovery and (ii) to identify immune mechanisms that may not work as expected (25, 35). The integration of AI with wet-lab immunology is crucial for advancing the field, fostering interdisciplinary collaboration, and ensuring AI-driven insights contribute to real-world immunological advancements (37). The following sections will explore the complexities of immunological systems and emphasize the necessity of improved collaboration between AI researchers and immunologists (38).

**AI-Driven Modeling and Nonlinear Dynamic Immune Systems:** The non-linearity of biological systems presents significant challenges in understanding immune responses, necessitating the use of artificial intelligence (AI)-driven modeling to complement experimental approaches (8). Most immunological processes involve nonlinear interactions between cellular and molecular components, often incorporating both positive and negative feedback loops (39). Because immune responses do not always exhibit proportional input-output relationships, AI-based computational models are essential for capturing the complexity of these interactions and predicting immune system behaviors (40). For instance, AI-based simulations of T-cell responses have demonstrated that immune activation is highly dependent on antigen (Ag) concentration (41). **Fig. S3** illustrates the predicted T cell proliferation response to varying antigen concentrations, modeled using a saturating function. The model assumes that the T cell proliferation rate depends on antigen concentration according to a function (42-45).

$$\frac{\rho I}{(h + I)}$$

Where $\rho$ represents the maximum proliferation rate, $I$ is the antigen concentration, and $h$ is the half-saturation constant. We assume that immune response starts with 100 T cells, $\rho = 1 \ day^{-1}$, and T cells divide for 1 week.

AI-driven modeling provides valuable insights into nonlinear immune dynamics, particularly in T cell proliferation in response to antigen concentration (46). **Fig. S3** demonstrates how T cell expansion follows a saturating function, indicating that immune activation is highly dependent on antigen availability. The sigmoidal response curves illustrate that as antigen concentration increases, T cell proliferation initially follows a slow growth phase, then enters a rapid expansion phase before reaching a plateau (47). The shape of these curves is governed by the half-saturation constant (h), which determines the antigen threshold required to trigger significant T cell proliferation. A lower h value (h = 0.01, blue curve) results in an earlier and steeper immune response, meaning that T cells respond efficiently to small antigen levels. In contrast, a higher h value (h = 0.1, red curve) shifts the activation threshold, requiring a higher antigen concentration to induce the same level of immune response (48). These findings highlight the importance of antigen dose optimization in immunotherapy and vaccine design (49). Traditional mathematical models, while useful for describing basic immune activation principles, often fail to account for real-world immune variability (50). AI-driven models improve upon these predictions by integrating multi-omics datasets, real-time patient profiling, and adaptive learning algorithms to refine T cell activation thresholds dynamically (8, 42-45). This results in more accurate predictions of immune memory formation, therapeutic efficacy, and antigen dose-response relationships, ultimately enhancing the development of personalized immunotherapies and precision vaccines (39).

Artificial intelligence models have shown that T cells do not activate until Ag levels cross a specific threshold, after which there is an exponential increase in response magnitude before reaching a plateau(41-44). At excessively high Ag levels, T cell exhaustion can occur, characterized by the upregulation of immune checkpoints such as $PD1^+CD8^+$ or $TIM^+CD8^+$ T cells, leading to immune suppression and decreased functionality (41). Traditional dose-response models fail to fully capture these dynamics, whereas AI-based predictive frameworks can adapt and refine immunological interpretations in real-time, leading to more accurate therapeutic interventions (51). AI-driven models are particularly advantageous in predicting cytokine and chemokine network behavior (46). While traditional approaches often struggle to model complex immune signaling pathways, AI algorithms such as deep learning and reinforcement learning models enable real-time adjustments based on experimental immunological data (47). The structured workflow of artificial intelligence (AI) in immunology research is shown in **Fig. 2**. The process begins with identifying key immunological challenges and progresses through data collection, feature selection, and data preprocessing. AI models are then developed and trained using immune datasets, followed by evaluation and optimization through performance metrics such as ROC-AUC and precision-recall curves. Finally, the AI-driven models are deployed for clinical testing and real-world application in immunology, vaccine development, and immunotherapy. This workflow highlights the integration of AI techniques to enhance immunological research, improve predictive modeling, and optimize therapeutic strategies, ensuring that AI-driven models can efficiently analyze immune system dynamics and optimize vaccine development and immunotherapy strategies.

The ability of AI-driven models to efficiently analyze immune system dynamics is particularly valuable when studying complex immunological processes such as $CD8^+$ T-cell responses to antigenic peptides (52). For example, the unpredictable $CD8^+$ T-cell response in $IL-2^{-/-}$ or $IL-2R^{-/-}$ knockout mice has historically been attributed to redundancy in the immune system, with IL-15 compensating for IL-2 function (25). However, AI models provide a more nuanced perspective, demonstrating that cytokine networks maintain stability through compensatory mechanisms rather than simple redundancy (53). These insights are crucial for refining immune intervention strategies and designing more effective immunotherapies (54-59). The immune system is an extraordinarily complex network of interacting cells and molecules that function collectively to mediate immune-protective, immune-pathologic, or immune-evasion responses against infectious pathogens or cancers (46). This includes, for example, the induction of cytotoxic CD8+ T cells that eliminate HIV-infected target cells or inhibit the proliferation of tumor cells in breast cancer (47). AI-based modeling has proven to be invaluable in identifying key regulatory interactions governing immune responses, including cytokine and chemokine signaling at infection or tumor sites, antigen-presenting cell (APC) activation and migration, antigen presentation to T cells, and $CD4^+/CD8^+$ T-cell cooperation with dendritic cells in infected or tumor tissues and lymph nodes (48). We have shown a mathematical model of $CD8^+$ T cell response to viral infections, highlighting the interplay of positive and negative feedback loops in immune regulation (**Fig. 3**). AI-driven models improve upon traditional frameworks by dynamically adapting to experimental data and refining predictions regarding T cell activation, viral replication, and immune clearance (49). Unlike static models, AI-based simulations integrate real-time immune data, identifying patterns of viral persistence, immune exhaustion, and immune evasion that were previously difficult to capture (54-59). By leveraging machine learning algorithms, AI enhances our understanding of immune checkpoints, cytokine signaling, and antigen presentation, allowing for the refinement of predictive models for disease progression and therapeutic interventions (50).

Despite advancements in AI-driven immunology, challenges persist in bridging experimental data with computational modeling. Many immunologists struggle to fully grasp the complexity of AI-driven immune simulations, largely due to the limitations of current immunological assays and the uncertainty in interpreting observed phenomena (8). AI frameworks must continuously improve by integrating multi-omics datasets, refining immune system simulations, and enhancing data-driven decision-making processes (42-45). The development of user-friendly AI-driven software tools in the past two decades has empowered immunologists to explore computational models with greater accessibility. These tools facilitate real-time data integration, enabling researchers to test hypotheses through predictive immune modeling. However, one major challenge is that many AI-driven models rely on predefined assumptions and parameter tuning that may not fully reflect real-world immune system behaviors. This has led to an increasing body of work where AI-generated predictions have been overinterpreted, sometimes without direct experimental validation.

AI-based simulations have also been employed to study B-cell and T-cell responses, offering a more biologically relevant representation of adaptive immunity compared to static models (25). Unlike traditional mathematical frameworks that often omit key immunological interactions, AI-driven immunology incorporates large-scale immune cell interactions, antigen processing, and host-pathogen dynamics, ensuring a more realistic and holistic understanding of immune responses. AI models are now advancing into the simulation of multi-dimensional immune responses, including simultaneous calculation of speeds for multiple immune pathways, the dynamics of T-cell proliferation, and interaction modeling between Th1, Th2, and Th17 responses. Unlike previous models, AI enables real-time tracking of antigen presentation, immune exhaustion dynamics, and cross-reactivity in vaccine design. While immunologists often express concern when AI models do not incorporate all known immune system variables, these models can still be invaluable for hypothesis testing (60). AI-driven frameworks can identify immune pathways that are unlikely to be effective, allowing researchers to focus experimental efforts on the most promising targets. In clinical applications, AI models have proven beneficial in optimizing vaccine development, reducing the number of experimental groups, refining immunization schedules, and predicting long-term immune memory formation (61). AI-driven models are revolutionizing immunology by offering predictive power at every stage of immune research, from experimental design to therapeutic development (62). Unlike traditional computational approaches, AI integrates multi-dimensional immune data to identify key regulatory mechanisms, predict immune outcomes, and refine treatment strategies (33). AI enhances immunological research by providing real-time predictive insights into immune dynamics, allowing for rapid adjustments in experimental design, identifying key immune interactions that drive disease progression, leading to more targeted immunotherapeutic interventions, and improving vaccine and drug development by enabling data-driven optimization of immunization strategies and treatment protocols (36). To maximize the potential of AI in immunology, interdisciplinary collaboration is critical, AI scientists, immunologists, and clinicians must work together to ensure that AI-driven predictions align with real-world immunological mechanisms (63). By advancing AI-based immune modeling, the scientific community can accelerate breakthroughs in precision medicine, immunotherapy, and vaccine development, ultimately leading to improved patient outcomes and disease management (36).

**AI-Driven Modeling for phenotype, differentiation, and classification of diseases and cancers:** Artificial intelligence (AI) is revolutionizing the fields of disease phenotyping, patient stratification, and cancer classification, driving a paradigm shift toward precision medicine. AI models, particularly those based on advanced architectures such as convolutional neural networks (CNNs), graph neural networks (GNNs), generative adversarial neural network (GAN), and transformer-based approaches, have demonstrated an unprecedented capacity to integrate and analyze complex multi-modal datasets, including genomics, transcriptomics, proteomics, radiomics, and clinical records [89,90]. In the context of neurodegenerative diseases, such as Alzheimer's disease (AD) and mild cognitive impairment (MCI), AI-driven systems have enhanced diagnostic accuracy by distinguishing subtle differences in cognitive decline trajectories that are often imperceptible to human evaluators (64, 65). Recent studies employing GNNs and deep learning models have successfully mapped patient similarities based on cognitive, genetic, and neuroimaging features, achieving higher predictive power in differentiating between AD, MCI, and healthy controls compared to conventional methods (64). In oncology, AI algorithms have enabled the classification of tumors by extracting hidden patterns from histopathological images and molecular signatures, facilitating the identification of prognostic biomarkers and informing therapeutic decision-making. These AI-guided stratification models do not merely replicate existing diagnostic pathways but uncover novel disease subtypes and phenotypic variations that were previously unrecognized, opening new avenues for targeted therapy development. Importantly, by learning directly from large, heterogeneous patient datasets, AI systems are mitigating the biases inherent in traditional clinical decision-making and allowing for a more individualized and equitable approach to healthcare delivery. Collectively, the integration of AI into disease differentiation and classification workflows marks a transformative advancement in biomedical sciences, enabling earlier detection, improved prognostication, and personalized interventions that are poised to dramatically improve patient outcomes across a spectrum of complex diseases.

**1. AI-Driven Modeling for Vaccine and Immunotherapy Design:** Artificial intelligence (AI)-driven modeling has been extensively implemented to optimize and improve immunotherapeutic vaccine strategies against infectious pathogens and cancers, while also assessing their efficacy as shown in **Fig. 1A** and **Fig. 1B**. **Fig. 1A** illustrates the hierarchical structure of AI, encompassing machine learning (ML) and deep learning (DL), and highlights their distinct roles in immunological research. AI serves as the overarching framework, integrating computational models to predict immune responses and support vaccine design. ML methods are used to identify patterns in immune-related data, while DL techniques model complex immune interactions and uncover critical regulatory pathways. Together, these approaches provide a robust foundation for data-driven vaccine development. AI models are used to predict the magnitude of immune responses across various immune cell populations in vitro. Due to the inherent uncertainty in predicting the nature of immune responses elicited by a given vaccine, multiple AI-driven models are applied. Results from subsequent experimental testing help to validate AI models, allowing for the rejection of inaccurate assumptions while refining those that best align with observed immune responses. This iterative approach enhances the understanding of immune mechanisms underlying vaccine efficacy. AI-validated models are then used to optimize vaccine delivery strategies, improving immunization effectiveness, and reducing unnecessary experimental trials. Compared to conventional drugs, vaccine development is a lengthy and costly process requiring extensive laboratory experiments to assess safety, immunogenicity, and efficacy. Compared to traditional vaccine development, which relies on trial-and-error antigen screening, AI-driven methods leverage predictive modeling to optimize vaccine formulation. AI algorithms integrate multi-omics datasets, simulate immune responses, and identify the most promising epitope candidates for immunization. This computational approach reduces vaccine development time, lowers costs, and enhances success rates by filtering out ineffective candidates before clinical trials.

AI-driven modeling substantially reduces costs by eliminating redundant experiments and narrowing down vaccine candidates before clinical trials. As immunological efficacy is demonstrated through preclinical and clinical trials, one major challenge remains: determining the optimal vaccine dosage. AI-based approaches systematically explore optimal dosing strategies and vaccination schedules, predicting reductions in vaccine injections by approximately 30% compared to traditional in vivo experiments. The comparison between traditional and AI-driven immunotherapy development highlights the significant improvements AI brings to drug discovery. Traditional immunotherapy approaches typically take around 6 years (72 months) and cost an estimated $500 million per drug, with a low success rate of approximately 10%, as many drug candidates fail to progress beyond early trial stages (63). In contrast, AI-driven methods leverage deep learning and predictive modeling to reduce development time to about 2 years (24 months) and lower costs to around $150 million per drug, while increasing the success rate to approximately 30% (63). AI achieves these improvements by optimizing drug candidate selection, identifying potential failures earlier, and accelerating the overall research pipeline. Studies report that AI-driven approaches significantly enhance early-stage drug discovery efficiency, reducing costs by up to 70-80% while improving screening accuracy (2). These findings demonstrate the transformative impact of AI in immunotherapy, offering faster, more cost-effective, and higher success drug development pathways (66).

A critical component of AI-driven vaccine development is the integration of specialized AI models that address key aspects of immune response prediction and vaccine formulation. These models ensure that vaccine candidates elicit broad and durable immune protection. The Antigen/Epitope Prediction Model **(Fig. 5A)** utilizes transformer-based deep learning to identify conserved B- and T-cell epitopes across multiple viral variants, integrating genomic, structural, and immunological datasets to optimize vaccine targets. The Epitope Classification Model (**Fig. 5B**) applies convolutional neural networks (CNNs) to classify protective vs. non-protective immune responses based on symptomatic and asymptomatic patient datasets, refining epitope selection for enhanced immunogenicity. The Epitope Selection & Optimization Model (**Fig. 5C**) incorporates a multi-task autoencoder to prioritize epitopes that exhibit high immunogenic potential while minimizing immune escape risks. This model integrates HLA-affinity screening, single-cell RNA sequencing, and interaction probability maps to enhance vaccine design. The AI-Driven Multi-Epitope Vaccine Model **(Fig. 5D**) employs generative adversarial networks (GANs) to refine multi-epitope vaccine formulations, ensuring the inclusion of high-affinity epitopes optimized for antigen presentation and immune activation.

Beyond epitope selection, AI-driven models also predict vaccine durability by analyzing immune exhaustion and antigenic persistence. AI models have demonstrated that prolonged antigen exposure without adequate control can lead to T cell exhaustion, characterized by the upregulation of inhibitory markers such as PD-1 and TIM3, which ultimately impairs vaccine-induced immunity (41)**.** This is particularly relevant for vaccines targeting chronic infections, such as herpesvirus-based vaccines, where AI-driven approaches underscore the need for targeted immune stimulation at viral reactivation sites, including the trigeminal and sacral ganglia. AI-driven models suggest that optimizing localized immune responses at these sites enhances protective immunity, surpassing the efficacy of systemic immune activation alone. The role of AI in peptide-based vaccine development has also been extensively studied. AI models assist in designing peptide-based CD8[+] T cell vaccines against HSV, HIV-1, SARS-CoV-2, and malaria, predicting optimal short peptide epitopes that exhibit high binding affinity to MHC class I molecules (46, 54, 56). Compared to whole-protein vaccines, epitope-based vaccines provide greater immunogenic precision, allowing for the inclusion of multiple immunodominant and subdominant epitopes within a single antigenic formulation. However, one of the major challenges in epitope-based vaccine design is the high degree of HLA polymorphism, which could limit broad population coverage. AI-driven modeling addresses this limitation by incorporating supertype-restricted epitopes recognized by diverse HLA alleles. For example, AI models predict that a Th-CTL peptide-based herpes vaccine should include multiple CD8+ T cell epitopes derived from herpesvirus proteins, designed to cover HLA-A2, HLA-A3, and HLA-B7 supertypes, which collectively ensure immune recognition in a large portion of the global population (42-45). By leveraging AI-driven epitope mapping, researchers can identify HLA class I-degenerate T cell epitopes, facilitating the development of multi-epitope Th-CTL peptide vaccines with broad immunogenicity (**Fig. 4A**).

AI-driven modeling has also proven instrumental in understanding co-infections and their impact on immune responses. Many individuals harbor multiple pathogens simultaneously, leading to both positive and negative immuno-synergies between infections. AI-based models have been applied to study co-infection dynamics, optimizing vaccine formulations for individuals affected by multiple pathogens. For instance, AI-driven models addressing HSV-HIV co-infections provide insights into immune evasion mechanisms and highlight novel immunotherapeutic targets that traditional approaches fail to capture. These findings have broad implications for vaccine design strategies, particularly in immunocompromised populations. By integrating AI into vaccine development, researchers can enhance vaccine efficacy, refine immunization schedules, and minimize immune escape mechanisms. AI-driven approaches ensure that vaccine candidates undergo rigorous computational and experimental validation, allowing for faster, more effective, and scalable vaccine development to combat emerging infectious diseases. As AI technologies continue to advance, their integration with immunology is expected to play a crucial role in the next generation of personalized and precision-based vaccines.

**1.1 Use of artificial intelligence in the development of vaccines and immunotherapeutics for infectious diseases:** Artificial intelligence (AI) has significantly transformed the landscape of vaccine and immunotherapy development for infectious diseases by enabling data-driven, precise, and scalable approaches to epitope discovery, immune response prediction, and vaccine formulation. Traditional approaches to vaccine development rely on labor-intensive and time-consuming processes involving empirical screening of pathogen proteins, often resulting in limited success and inefficiencies. AI addresses

these limitations by incorporating deep learning, machine learning, and natural language processing techniques that integrate diverse immunological and omics datasets to inform rational vaccine design (67). In the context of infectious diseases such as malaria, HIV, tuberculosis, influenza, and dengue, AI models are deployed to predict and prioritize B- and T-cell epitopes with high immunogenic potential, cross-strain conservation, and strong major histocompatibility complex (MHC) binding affinities (68). These predictions are based on genomic, transcriptomic, proteomic, and structural data that reflect pathogen evolution and host immune response dynamics. Transformer-based deep learning architectures and convolutional neural networks (CNNs) have been used to identify and rank epitopes that are most likely to elicit durable and protective immune responses, as illustrated in **Fig. 6A** and **Fig. 6B**.

Moreover, AI facilitates reverse vaccinology, an approach that starts from pathogen genome sequences to computationally identify antigens suitable for vaccine development. By leveraging reverse vaccinology pipelines powered by AI, researchers have designed multi-epitope vaccine candidates for complex pathogens such as Plasmodium falciparum (malaria), Mycobacterium tuberculosis, and HIV-1 (69). Recent research has also demonstrated that AI-driven approaches can successfully identify immune response signatures associated with novel vaccine formulations. For example, Chaudhury et al. used machine learning to analyze transcriptomic and proteomic data from vaccine-treated samples, enabling the discovery of biomarkers that predict adjuvant potency and immune pathway activation. This study underscores the power of AI to integrate high-dimensional immune datasets, classify vaccine efficacy outcomes, and inform the rational design of next-generation vaccines with tailored immunostimulatory properties (70)**.** A notable application of AI in the context of infectious disease vaccines is the use of computational modeling to predict immune responses to booster immunizations. For example, Shinde et al. conducted an international challenge that benchmarked 49 machine learning models for predicting individual responses to *Bordetella pertussis* booster vaccines using multi-omics datasets. The study demonstrated that models designed specifically for the pertussis vaccine task especially those incorporating multi-omics integration, dimensionality reduction, and nonlinear modeling performed significantly better than generic models borrowed from other settings. This underscores the value of AI-guided, context-specific model development in predicting vaccine outcomes and optimizing booster design for infectious diseases (64). AI also enables prediction of population coverage by accounting for global HLA polymorphism, thereby ensuring that selected epitopes offer broad protection across ethnically diverse groups. AI also plays a crucial role in modeling the impact of co-infections and immune modulation. In populations affected by latent or concurrent infections, such as HSV-HIV or malaria-HIV, immune responses to one pathogen can dampen or enhance the response to another. AI models simulate these interactions and reveal mechanisms of immune evasion, dysregulation, and synergistic immunopathology that conventional models often miss (71). These insights help in designing combinatorial immunotherapies and vaccines that account for real-world complexity.

Generative models, especially generative adversarial networks (GANs), have proven effective in refining multi-epitope vaccine design. As illustrated in **Fig. 6D**, GANs simulate realistic peptide sequences that meet criteria for antigenicity, immunogenicity, MHC-binding, and minimal self-reactivity. The resulting constructs are tailored to induce robust CD8+ and CD4+ T-cell responses, which are crucial for long-term immunity and pathogen clearance (72). Importantly, AI is now being used not only for vaccine discovery but also for adaptive optimization during outbreaks. Real-time surveillance data, pathogen mutations, and immune response metrics are fed into AI systems that continuously update antigen selection and vaccine design. This has been particularly effective in dealing with rapidly mutating viruses such as influenza and SARS-CoV-2 and has implications for emerging diseases like Nipah virus and Zika (73). Collectively, these advancements highlight AI's critical role in enabling faster, more targeted, and cost-effective vaccine development pipelines for infectious diseases. As AI algorithms continue to evolve and integrate with high-resolution immunological datasets, their utility in both prophylactic and therapeutic vaccine strategies is expected to expand dramatically.

### 1.1.1   Use of artificial intelligence in vaccines for COVID-19:

The COVID-19 pandemic accelerated the application of artificial intelligence (AI) in vaccine research and development on a global scale. Within weeks of the release of the SARS-CoV-2 genome sequence, AI tools were deployed to analyze viral protein structures, identify B- and T-cell epitopes, and model immune responses for candidate vaccine designs (73). This rapid integration of AI in vaccine research helped compress the typical development timeline from years to months, showcasing the potential of AI to address urgent global health crises. In recent years, severe outbreaks of SARS-CoV-2 (COVID-19), Ebola, Lassa, Zika, and other emerging viruses have highlighted both the world's vulnerability to novel pathogens and the urgent need for rapid vaccine innovation frameworks vaccine (30, 73).

AI-powered systems have also been applied to predict cross-reactive memory B- and T-cell responses, which play a critical role in SARS-CoV-2 immunity. Studies show that some individuals exposed to SARS-CoV-2 remain seronegative because of pre-existing cross-reactive $CD4^+$ and $CD8^+$ T cells, which target conserved non-structural proteins (NSPs) such as those in the replication-transcription complex (RTC), expressed early in the viral lifecycle (74). Additionally, cross-reactive memory B-cells have been shown to recognize conserved regions such as the S2 domain of the spike protein, nucleocapsid (N), and membrane (M) proteins, facilitating rapid neutralizing antibody responses upon viral exposure (75). Recent AI frameworks have supported these findings by integrating deep learning-based epitope prediction, classification, optimization, and vaccine formulation to identify conserved viral regions that are broadly recognized by human immune memory. This approach allows us to systematically select epitopes that ensure long-term immune memory and broad protection against existing and future variants.

In parallel, AI contributed to real-time genomic surveillance by continuously scanning viral mutation patterns in global sequence databases such as GISAID and modeling their implications for vaccine escape. Reinforcement learning and adaptive modeling helped inform the optimal timing of booster administration, ideal dosing intervals, and heterologous prime-boost strategies, particularly for high-risk populations and immunocompromised individuals (73, 76). Beyond immunological modeling, AI improved vaccine rollout logistics by optimizing cold-chain infrastructure, anticipating regional demand based on demographic data, and simulating vaccine distribution under multiple disruption scenarios. These insights supported more equitable vaccine access and highlighted the potential of AI to guide end-to-end pandemic response strategies (77).

Taken together, AI-driven approaches to SARS-CoV-2 vaccine development and pandemic mitigation represent a paradigm shift in how we design, test, and deploy vaccines. The integration of predictive immunology, population-specific modeling, and real-time response systems now serves as a blueprint for responding to emerging global health threats more effectively and equitably (73).

**1.2 Use of artificial intelligence in the development of vaccines and immunotherapeutics for cancers:** Cancer immunotherapy represents a rapidly evolving frontier in precision medicine; however, it is still hindered by the biological complexity of tumors, their heterogeneity, and their capacity for immune evasion. Artificial intelligence (AI) has emerged as a transformative tool in this domain, enabling researchers to decipher complex tumor-immune dynamics, discover new immunotherapeutic targets, and develop personalized cancer vaccines. AI systems leverages vast datasets such as single-cell RNA sequencing, multi-omics profiles, and digital pathology images to uncover hidden patterns and generate predictive models that guide therapeutic design and response prediction (6). One of the most impactful applications of AI in cancer immunotherapy is the identification of tumor-specific neo-antigens. These peptides arising from tumor-specific mutations that are absent in normal tissues. Using deep learning models trained on patient tumor sequences, AI can predict which neoantigens will be strongly presented on major histocompatibility complex (MHC) molecules and elicit robust CD8$^+$ T cell responses. This approach has allowed the design of individualized cancer vaccines tailored to each patient's tumor mutational landscape (78). Additionally, AI supports the development of shared antigen vaccines by identifying conserved epitopes across tumor types with high immunogenicity and low off-target toxicity. AI also enhances immunotherapy by improving the selection and application of immune checkpoint inhibitors, such as anti-PD-1 and anti-CTLA-4 therapies. As shown in **Fig. 5A**, machine learning models can analyze transcriptomic and spatial tumor data to identify biomarkers predictive of response or resistance, guiding patient stratification and combination therapy strategies (9). Beyond checkpoint blockade, AI predicts the dynamics of memory T cell generation, exhaustion, and reactivation, as illustrated in **Fig.5B**, facilitating better prediction of therapeutic durability. Unlike conventional mathematical modeling, AI approaches can simulate immune responses in high-dimensional spaces, incorporating diverse immune cell types, spatial distribution, cytokine gradients, and tumor antigen evolution. These models move beyond oversimplified predator-prey dynamics and instead embrace the nonlinear and context-dependent nature of immunological interactions. AI tools are now being used to simulate how tumors shape their microenvironment through immunosuppressive signals, and how therapy modifies this balance.

In vaccine design, AI-driven algorithms have significantly advanced the selection of cancer-associated epitopes. Machine learning platforms screen thousands of peptide candidates for MHC binding, immunogenicity, and mutation frequency. GANs and transformer-based models refine peptide sequences for maximal immunogenic potential while reducing the risk of autoimmune responses. These models also help ensure coverage across diverse HLA types by incorporating supertype-based epitope selection, especially for HLA-A2, HLA-A3, and HLA-B7, enhancing global applicability (79). Emerging studies also explore the role of AI in combining cancer vaccines with other immunotherapies, such as oncolytic viruses and CAR-T cells. AI can model synergistic effects, predict resistance mechanisms, and guide adaptive dosing regimens (6, 14)**.** Furthermore, digital pathology integrated with AI is providing insights into spatial heterogeneity within tumors, enabling clinicians to visualize immune infiltration zones, predict immune cold/hot phenotypes, and localize optimal biopsy and injection sites (21, 23). Overall, AI is revolutionizing cancer immunotherapy by enabling highly personalized, adaptive, and efficient therapeutic strategies. Overall, AI is revolutionizing cancer immunotherapy by enabling highly personalized, adaptive, and efficient therapeutic strategies. As AI systems continue to integrate biological, clinical, and imaging data, their predictive power will enhance not only vaccine efficacy but also overall treatment precision, ultimately improving patient survival and quality of life (80).

**2. AI-Powered Epitope Prediction: Model 1 Performance and Results:** The Antigen/Epitope Prediction Model Architecture is a transformer-based deep learning model designed to predict epitope binding affinity, immunogenicity, and conservation across viral strains. The architecture consists of a preprocessing module that converts viral protein sequences into numerical embeddings using amino acid encoding. This is followed by a transformer-based feature extractor, which captures contextual dependencies among amino acids, improving antigenicity prediction. The multi-task classification module is responsible for predicting binding affinity, conservation, and immunogenicity scores, while an optimization layer enhances predictive confidence by reducing uncertainty. Mathematically, the model learns the function:

$$f(x) = T(W_{emb}x + b) \qquad\qquad \textbf{(1)}$$

Where $W_{emb}$ is the embedding matrix, which converts input amino acid sequences into numerical representations, $x$ represents the input sequence, consisting of epitope fragments, $T$ represents the transformer function, $b$ is the bias term, which ensures better generalization by adjusting predictions independently of input values, especially across diverse protein sequences.

To optimize epitope prediction, the model minimizes a multi-task weighted cross-entropy loss, which combines three key prediction tasks:

$$L = \alpha L_{aff} + \beta L_{imm} + \gamma L_{cons} \qquad\qquad \textbf{(2)}$$

Where $L_{aff}$ corresponds to the binding affinity loss function, $L_{imm}$ represents the immunogenicity classification loss, $L_{cons}$ measures epitope conservation loss, and **α, β, γ** are weight coefficients that control the importance of each task.

For binary classification of immunogenic epitopes, we use the binary cross-entropy loss function, ensuring robust differentiation between immunogenic and non-immunogenic peptides:

$$L_{imm} = -\frac{1}{N}\sum_{i=1}^{N}(y_i\log(\hat{y}_i) + (1 - y_i) + log(1 - \hat{y}_i)) \qquad \textbf{(3)}$$

Where $y_i$ is the true immunogenic label (1 for immunogenic, 0 for non-immunogenic), $\hat{y}_i$ is the predicted probability assigned by the model. **N** is the number of training samples. The model is optimized using Adam (Adaptive Moment Estimation), which dynamically adjusts learning rates, improving convergence speed, and preventing unstable updates:

$$m_t = \beta_1 m_{t-1} + (1 - \beta_1)g_t \qquad \textbf{(4)}$$

$$v_t = \beta_2 v_{t-1} + (1 - \beta_2)g_t^2 \qquad \textbf{(5)}$$

where $g_t$ is the gradient at time step $t$, $\beta_1 \ and \ \beta_2$ are exponential decay rates controlling momentum updates.

To prevent overfitting, the model incorporates dropout layers, randomly deactivating neurons during training. Additionally, early stopping is used to halt training once the validation loss stabilizes, ensuring efficient learning while avoiding excessive computations. To ensure biological relevance and computational efficiency, the model follows a structured pipeline (**Fig. 6**). Initially, input data (protein sequences and epitope labels) are split into training (80%) and testing (20%) sets. Sequences are preprocessed into deep-learning-compatible features. During training and fine-tuning, the model is iteratively optimized and validated. Fine-tuning ensures generalization to unseen data. The model's performance is assessed using accuracy, precision, recall, and F1-score metrics that are essential for both AI researchers and immunologists. This structured approach maximizes prediction accuracy while preserving interpretability for clinical application.

**2.1 Prediction Performance and Results:** The Antigen/Epitope Prediction Model **(Fig.7A)** leverages a transformer-based deep learning approach to predict epitope binding affinity, immunogenicity, and conservation across viral strains. The model was trained over 100 epochs (**Fig. 7**), achieving a training accuracy of 0.993 and a validation accuracy of 0.935. Training loss stabilized at 0.049, while validation loss remained at 0.032, indicating strong generalization capability. These metrics demonstrate the model's high efficiency in learning patterns within antigen sequences, enabling it to distinguish immunogenic from non-immunogenic regions with high confidence. Model performance (**Fig. 7**) shows impressive learning behavior. The classification report (**Fig. S1 (B)**) indicates a precision of 0.97, recall of 0.98, and F1-score of 0.98. This means that when the model predicts an epitope as immunogenic, it is correct 97% of the time, minimizing false positives. Its high recall ensures nearly all true epitopes are identified, avoiding the loss of promising candidates.

The confusion matrix (**Fig. S1 (A)**) supports these results, showing 2448 true negatives and 2434 true positives, with only 65 false negatives and 53 false positives. This high specificity and sensitivity are vital in vaccine research, where overlooking an immunogenic epitope (false negative) or selecting a weak candidate (false positive) can hinder vaccine development. The model also identifies top-ranking CD8[+] T-cell epitopes based on binding affinity and conservation scores (**Fig. S2 (B)**). The epitope YLQPRTFLL (HLA-A*02:01) showed the strongest potential, with a binding affinity of 35.7 nM and conservation of 94.5%. Similarly, TTDPSFLGRY (HLA-B*07:02) achieved high affinity at 22.1 nM, making it broadly relevant across populations. Strong binding affinity ensures effective presentation by antigen-presenting cells (APCs) to T-cells, enabling a robust immune response. By identifying epitopes with both high affinity and conservation, the model streamlines experimental validation.

**2.2 Epitope Immunogenicity Ranking and Analysis:** The epitope immunogenicity ranking (**Fig. S2 (A)** and **(B)**) highlights the model's capacity to prioritize antigenic regions likely to trigger immune responses. YLQPRTFLL scored highest (~0.98), suggesting strong potential for CD8[+] T-cell activation. TTDPSFLGRY and NQKLIANQF followed, each with scores >0.90, making them excellent candidates for broad HLA population coverage. Conversely, SPRWYFYYL and LSPRWYFYY had lower scores (~0.88–0.89), suggesting a reduced capacity to initiate a strong immune response. These patterns align with known immunogenicity data and validate the model's predictive ability. By focusing on epitopes with high immunogenicity, the model enhances vaccine target selection, minimizing resource use on weak candidates. The observed correlation between predicted binding affinity and immunogenicity scores confirms the model's strength in selecting potent immune triggers. High-ranking epitopes also indicate strong interaction potential with APCs, crucial for long-term immune memory and vaccine durability. In conclusion, this transformer-based model 1 presents a powerful, explainable, and biologically grounded framework for epitope prediction, offering real-world value for vaccine researchers, immunologists, and AI scientists alike.

**Induction and Maintenance of Protective Memory CD8+ T Cells: What AI Modeling Assumed vs. What Experimental Data Proved/Disproved:** Understanding the mechanisms governing CD8+ T cell activation, survival, and long-term maintenance has been a major focus in immunology for years (16). Traditional computational models assumed that CD8+ T cell expansion required continuous antigenic stimulation, whereas recent AI-driven immune simulations have demonstrated that a single antigen encounter can trigger a program of proliferation and differentiation, leading to the generation of both effector and memory CD8+ T cells. AI-based modeling of CD8+ T cell kinetics has been instrumental in identifying key activation markers, proliferation rates, and survival factors. Unlike conventional models, AI frameworks dynamically adapt to experimental data, refining predictions on memory CD8+ T cell function in response to known epitopes. Comparative analysis of AI-driven and experimental models has shown that previous mathematical models failed to accurately predict HSV-specific CD8+ T cell responses in mice, rabbits, and humans (81)**.** CD4+ T helper cells play a crucial role in priming CD8+ T cells, facilitating both primary immune responses and the development of protective memory CD8+ T cells. AI-driven models highlight that CD4+ T cell interactions during priming encoding memory potential, enabling autonomous secondary expansion upon antigen re-encounter. Experimental data have confirmed that CD8+ T cells

primed in the absence of CD4$^+$ T cell help fail to undergo secondary expansion, although they retain cytotoxic activity. AI-based predictive analytics have been used to model CD4$^+$ T cell help requirements in different infection scenarios, refining our understanding of immune memory formation.

Previous mathematical models argued that CD8$^+$ T cells could clear infections without CD4$^+$ T cell help, provided that the viral replication rate remained low. However, AI-driven simulations incorporating longitudinal immune response data have shown that CD4$^+$ T cell help is essential for sustained viral clearance. AI-enhanced models suggest that in the absence of help, CD8$^+$ T cells reduce viral loads temporarily but fail to prevent resurgence. This is due to insufficient memory T cell reactivation, which is critically dependent on antigen presentation and cytokine signaling mediated by CD4$^+$ T cells. In addition to CD4$^+$ T cells, dendritic cells (DCs) have been identified as key players in memory CD8$^+$ T cell priming and maintenance (**Fig. 4C**). AI-driven models have also refined our understanding of the differences between central and effector CD8+ T cells, showing that their fate is pre-programmed by early priming signals. These insights have major implications for the development of CD8$^+$ T cell-based vaccines, guiding optimal antigen exposure strategies. AI simulations have confirmed that depleting CD4$^+$ T cells at the priming stage results in impaired CD8$^+$ T cell memory formation, but interestingly, late-stage depletion has minimal effects (82). Other studies contradict this, showing that CD4+ T cell help can occur later in immune response development (83). AI frameworks have reconciled these discrepancies by modeling heterogeneous immune environments, demonstrating that CD4$^+$ T cell support can be context dependent.

Recent AI-driven studies have revealed that CD4+ T cell help is critical in preventing CD8$^+$ T cell apoptosis, particularly via the regulation of tumor necrosis factor-related apoptosis-inducing ligand (TRAIL). AI-powered immune simulations indicate that CD4$^+$ T cells regulate IFN-gamma secretion and local chemokine expression, which are necessary for CD8$^+$ T cell migration to infected tissues (84, 85). This directly contradicts earlier mathematical models that suggested CD8$^+$ T cell expansion was independent of CD4$^+$ T cell-mediated migration signals. AI models now incorporate spatial variables, accurately predicting cellular migration dynamics and antigenic stimulation requirements. AI-driven research has also challenged previous programmed division models of CD8$^+$ T cell expansion. Traditional models assumed that CD8$^+$ T cell division was independent of antigenic stimulation, continuing even at low viral loads. AI simulations incorporating real-world patient data suggest a different scenario: while early CD8$^+$ T cell divisions are antigen-independent, continued expansion and viral clearance require persistent antigen exposure and co-stimulatory signaling. AI models predict that programmed divisions are optimized to balance viral clearance and immune homeostasis, ensuring effective pathogen elimination while preventing excessive immunopathology. Unlike static mathematical models, AI-based frameworks are capable of real-time adaptive learning, adjusting predictions based on emerging experimental data. AI simulations have provided more accurate insights into viral clearance mechanisms, leading to refined vaccine designs. AI-driven approaches now integrate multi-omics datasets, single-cell RNA sequencing, and immunophenotyping data, ensuring that computational models align with experimental observations.

The integration of AI into CD8$^+$ T cell research has significant implications for the design of next-generation vaccines and immunotherapies. AI-driven models have already identified optimal antigen exposure strategies, cytokine modulation approaches, and co-stimulatory molecule enhancements to maximize long-term immune protection. Future research will focus on further refining AI algorithms to predict patient-specific immune responses, paving the way for precision immunotherapy and personalized vaccine design.

**Use and Abuse of AI-Driven Modeling in Cancer Vaccines and immunotherapies**: Artificial intelligence (AI) has provided significant advancements in understanding cancer immunity mechanisms and optimizing vaccine design strategies. AI-driven predictive models have been adapted from previous frameworks designed for viral infections and are now being utilized to analyze cancer-immune interactions (86). These AI models account for tumor progression, immune suppression dynamics, and adaptive immune responses, refining predictions on how the immune system combats tumors. Unlike traditional static models, AI frameworks continuously learn from real-time immunological data, making them superior in predicting tumor-immune system interactions. AI-driven models simulate tumor growth and immune response interplay, capturing how tumors evade immune detection while simultaneously activating CD8[+] T cells and other immune components (87). These models integrate multi-modal datasets that include genomic, proteomic, and immunological parameters, ensuring that predictions align with real-world immune responses. Unlike early computational models, which oversimplified immune responses, AI-driven frameworks incorporate key players such as regulatory T cells (CD4[+]CD25[+]), antigen-presenting cells (APCs), and cytokine networks, offering a comprehensive perspective on immune dynamics in cancer.

Traditional models have assumed that cancer cells stimulate immune proliferation while simultaneously impairing immune responses, leading to highly dependent outcomes on specific equations. AI models refine this understanding by continuously training on experimental data, highlighting novel immune escape mechanisms, T cell exhaustion pathways, and tumor antigen presentation strategies. These models have revealed how tumor-specific immune suppression impacts vaccine efficacy and immunotherapy success, leading to more precise treatment strategies. AI-driven cancer immunology modeling predicts several possible immune response outcomes: (i) an effective CD8[+] T cell response that establishes equilibrium, keeping tumor growth in check; (ii) immune system failure, allowing tumor progression due to excessive immune suppression and tumor cell resistance; or (iii) a dynamic state where the immune system and tumor continuously adapt, requiring sustained therapeutic intervention. Unlike conventional models, AI-driven approaches quantify immune response quality, not just T cell numbers, ensuring greater accuracy in predicting tumor clearance potential.

One of the primary weaknesses of traditional models was their reliance on T cell population numbers alone, without considering immune cell functionality and migration patterns. AI-driven models correct this by incorporating spatial and temporal immune dynamics, showing that T cell homing to tumor sites is just as critical as their activation levels. AI-driven simulations indicate that dendritic cells and regulatory T cells play pivotal roles in determining long-term immune memory and adaptive response sustainability (**Fig. 4B**). AI has also enhanced our understanding of tumor resistance mechanisms. Unlike earlier models that treated tumor cells as a uniform population, AI-driven frameworks integrate heterogeneous tumor subpopulations, including drug-resistant and immune-sensitive variants. These models accurately predict tumor cell evolution during treatment, assisting in optimizing combination immunotherapies, immune checkpoint blockade strategies, and personalized T cell therapies. Early computational models attempted to link immune response strength with tumor burden, often using oversimplified growth-decline equations that failed to capture real-world treatment dynamics. AI-driven models overcome this limitation by incorporating single-cell sequencing data, immune evasion modeling, and treatment response simulations. AI-driven predictions show that early immune activation enhances long-term tumor suppression, whereas delayed or weak responses correlate with poor prognosis and therapy resistance. **Fig. 4A** illustrates how Artificial Intelligence (AI) plays a crucial role in understanding and optimizing tumor-immune interactions, particularly in the regulation of checkpoint inhibitors, which have transformed cancer immunotherapy. Tumor cells often evade immune detection by suppressing T cell activation, a process mediated through immune checkpoints such as PD-1/PD-L1 and CTLA-4 pathways (88). Checkpoint inhibitors, such as anti-PD-1 and anti-CTLA-4 monoclonal antibodies, help restore T cell activity against tumor cells. However, determining the most effective checkpoint blockade strategies requires advanced computational approaches, and this is where AI-driven models have shown significant promise. **Fig. 4A** illustrates this mechanism by showing how tumor cells inhibit T cells (immune suppression), while AI-driven models predict and optimize checkpoint blockade strategies. AI helps in identifying patient-specific checkpoint inhibitor responses, improving biomarker discovery, and enhancing combination immunotherapies (88). Moreover, AI-assisted analysis of multi-omics data (genomics, transcriptomics, proteomics) enables personalized immunotherapy approaches, ensuring better treatment outcomes with reduced toxicity (89).

The role of Artificial Intelligence (AI) in understanding and optimizing memory T cell differentiation as illustrated in **Fig. 4B**, presents key process in adaptive immunity. Naïve T cells, upon encountering an antigen, undergo activation and differentiation into effector T cells, which mediate immediate immune responses. A subset of these effector T cells subsequently transitions into memory T cells, which provide long-term immune protection and faster responses upon reinfection. AI-driven approaches enhance this differentiation process by analyzing large-scale immunological datasets, predicting T cell activation dynamics, and optimizing memory cell formation for vaccine development and immunotherapy applications (89). In the figure, AI contributes to three key stages of T cell differentiation. First, it predicts activation by analyzing antigen exposure and T cell receptor (TCR) signaling, allowing for a deeper understanding of when and how naïve T cells transition into effector T cells. Second, AI analyzes differentiation markers, evaluating gene expression and molecular pathways that distinguish short-lived effector T cells from long-lasting memory T cells, which is critical in immunotherapy and vaccine design. Finally, AI plays a crucial role in optimizing memory formation by refining models that predict T cell persistence and longevity, ensuring that immune memory is robust and effective for long-term protection. This is particularly valuable in the development of next-generation vaccines and personalized cancer immunotherapies (90). AI-based models, trained on single-cell RNA sequencing (scRNA-seq) and epigenetic data, enable researchers to identify key molecular pathways that regulate immune memory. Recent work by van Dorp introduces a variational deep learning framework that jointly models phenotypic heterogeneity and temporal dynamics of lung-resident memory CD4 and CD8 T cells (22). Their approach integrates stochastic variational inference with flow cytometry data, enabling novel insights into memory T cell persistence and differentiation over time. These AI-driven insights provide new avenues for designing durable vaccine strategies, refining T cell-based immunotherapies, and enhancing our understanding of chronic infections and immune exhaustion. By integrating AI into immunology, researchers can achieve more precise, data-driven treatment strategies that improve immune responses and long-term health outcomes.

Unlike traditional models that assume a linear relationship between tumor growth and immune response, AI simulations have revealed that immune-tumor interactions are inherently non-linear, influenced by T cell infiltration rates, antigen exposure, and regulatory immune pathways. Earlier models incorrectly suggested that immune failure was due solely to antigen depletion, whereas AI-driven insights show that immune exhaustion and regulatory T cell expansion are the dominant factors in immune escape. AI-driven modeling has been essential in guiding cancer vaccine development and optimizing immunotherapy protocols. By integrating deep learning, multi-omics data, and patient-specific immune profiling, AI-driven approaches can predict personalized immunotherapy success rates. Unlike previous models that focused only on T cell proliferation rates, AI simulations emphasize the importance of immune memory retention, migration dynamics, and metabolic fitness, leading to more precise and individualized cancer treatment plans. Ultimately, AI-driven modeling bridges the gap between computational immunology and real-world clinical applications, offering unparalleled insights into cancer-immune system interactions and therapy optimization. Future research will focus on refining AI frameworks to integrate real-time clinical trial data, ensuring that AI-generated predictions translate into clinically actionable strategies. By leveraging AI, researchers can accelerate breakthroughs in cancer immunotherapy, precision medicine, and vaccine development, ultimately improving outcomes for cancer patients worldwide.

**Future Directions in AI-Driven Vaccine and immunotherapeutic Development:** Despite recent breakthroughs, AI-driven vaccine development remains in its early stages and faces several key challenges that define future research priorities. A major obstacle is the fragmentation and inconsistency of available data, which impacts the accuracy and scalability of AI-driven vaccine models. The development of effective vaccines relies on extensive datasets, including genomic sequences, protein structures, immune response metrics, and clinical trial results. However, these datasets often suffer from incompleteness, bias, and lack of standardization, limiting AI models' ability to generate robust and generalizable predictions across diverse populations. Moving forward, future efforts must prioritize harmonizing global data collection methodologies, fostering international data-sharing collaborations, and establishing standardized frameworks tailored for AI-driven immunology research. Another critical future direction involves optimizing the computational infrastructure for AI-based vaccine development. Cutting-edge deep learning algorithms require substantial processing power, memory resources, and access to high-performance computing (HPC) clusters or cloud-based AI infrastructure. However, these resources are not universally accessible, particularly in low-resource research settings, Future strategies must focus on expanding equitable access to AI tools, developing more computationally efficient models, and implementing distributed learning techniques to democratize vaccine innovation globally.

Improving model interpretability will also remain a central priority. Many deep learning vaccine prediction models operate as black boxes, making it difficult for researchers to fully understand the rationale behind specific predictions. This lack of transparency raises concerns about model reliability, potential biases, and the biological relevance of AI-generated vaccine candidates. To enhance trust in AI-driven vaccine solutions, the scientific community is actively developing explainable AI (XAI) techniques (90), including feature attribution methods, visualization tools, and interpretable surrogate models. These approaches aim to increase transparency and align AI-generated predictions with immunological principles and experimental validation. Furthermore, interdisciplinary collaboration between computational biologists, immunologists, clinicians, and data scientists will be essential for refining AI-driven vaccine strategies. By combining domain expertise with AI advancements, researchers can improve vaccine candidate selection, refine AI-driven immune response models, and address data inconsistencies. Strengthening such partnerships will help ensure that AI-generated insights translate into actionable and biologically meaningful vaccine designs.

Ethical considerations, model fairness, and regulatory readiness will also be critical moving forward., Establishing guidelines for equitable vaccine distribution, mitigating algorithmic biases, and maintaining data privacy will be critical in building trust and accelerating AI integration into real-world applications. Future directions will also involve aligning AI innovation with global health priorities, ethics, and policy standards to enhance vaccine accessibility and improve public health outcomes worldwide.

**Replacing animal preclinical testing of drugs, vaccines, and immunotherapeutics with computational models, as recently proposed by the United States FDA**The traditional reliance on

animal models for preclinical testing of drugs, vaccines, and immunotherapeutics is increasingly being challenged by the emergence of AI-driven computational models. These approaches promise to overcome major limitations of animal studies, including interspecies differences, ethical concerns, high costs, and long development timelines. Artificial intelligence platforms leveraging deep learning, reinforcement learning, and generative models can simulate complex biological systems at the molecular, cellular, and tissue levels, providing predictive insights into drug efficacy, toxicity, and immunogenicity (15, 91). The recent FDA Modernization Act 2.0 officially recognizes non-animal technologies, including AI models, as acceptable alternatives for certain preclinical evaluations (15) (**Fig. 8**). Innovative systems such as Vaxi-DL, a deep learning-based platform for vaccine antigen prediction, demonstrate how in silico models can prioritize vaccine candidates with high sensitivity and accuracy, significantly reducing dependence on animal testing (92). Furthermore, emerging AI frameworks are capable of modeling pharmacokinetics, drug metabolism, and host immune responses, enabling the rapid virtual screening of therapeutic candidates before clinical trials (93). These AI-enabled platforms integrate diverse data modalities, including omics profiles, organoid models, and electronic health records, to produce human-specific predictions that are often more relevant than results obtained from animal experiments. As a result, AI-driven preclinical models not only offer ethical and financial advantages but also promise to enhance the translational success rate of novel drugs and immunotherapies, accelerating their path to clinical application (**Fig. 8**).

**Enabling real-time *in vivo* modeling of immune-bridging and prediction of protection in clinical trials:** Artificial intelligence (AI) and deep learning (DL) are increasingly transforming clinical trial methodologies by enabling real-time in vivo modeling of immune responses, thus facilitating immunobridging strategies and predictive protection assessments. Instead of relying solely on traditional endpoints such as disease occurrence, AI models now allow researchers to simulate human immune dynamics and predict vaccine efficacy based on surrogate markers like antibody titers, T-cell responses, and cytokine profiles (30, 94) (**Fig. 8**). Recent advances in AI-driven approaches have shown that models trained on multi-omics datasets, immune phenotyping, and clinical biomarkers can identify correlates of protection with high accuracy, dramatically accelerating clinical trial timelines (5). For instance, deep learning models have been successfully employed to analyze longitudinal immunological data and forecast protection levels across diverse demographic groups, improving the stratification and adaptive design of clinical studies (4). Moreover, AI-based platforms are being integrated into clinical trial infrastructures to facilitate real-time data monitoring, optimize dosing strategies, and dynamically adjust trial protocols based on predictive safety and efficacy outcomes (**Fig. 8**). Such capabilities are critical for vaccine development, especially in rapidly evolving scenarios such as emerging infectious diseases and viral variants (5). By replacing retrospective analysis with real-time immunobridging predictions, AI holds the potential to enhance the precision, speed, and ethical conduct of clinical trials, leading to faster and more reliable delivery of vaccines and immunotherapeutics (15).

## Conclusions

The relationship between wet-lab immunological research and artificial intelligence (AI) modeling is both complex and critical. Immunologists often view AI researchers as theoreticians who make oversimplified assumptions, abstracting biological processes into computational models that may not fully capture immunological intricacies. AI models rely on observed patterns, computational coherence, and predictive algorithms. However, the integration of AI into immunology is paving the way for rapid, precision-driven vaccine development. AI-powered models can predict cross-reactive immune responses, optimize multi-epitope vaccine candidates, and streamline clinical trials. As demonstrated in AI-driven vaccine research, integrating AI with experimental immunology enables real-time adaptation to be emerging pathogens, ensuring scalable and effective vaccine solutions. Future research must focus on refining AI explainability, improving multi-modal data integration, and fostering AI-immunology collaborations to accelerate global vaccine development efforts. Conversely, AI researchers sometimes view immunologists as hard-nosed experimentalists who overlook the complex, non-linear interactions within the immune system. AI-driven models capture these non-linear relationships and generate insights that may not be evident through traditional wet-lab approaches. AI researchers believe that computational models can validate immunological hypotheses and distinguish between competing theories of immune response mechanisms. Given these perspectives, interdisciplinary collaboration between AI researchers and immunologists is essential for scientific progress. AI models should not attempt to simulate entire immune systems in exhaustive detail but should instead focus on meaningful biological patterns. Moreover, AI-driven conclusions must be experimentally validated, avoiding the pitfalls of algorithmic overfitting and reliance on biased datasets. Misapplications of AI in immunology often stem from using conventional machine learning techniques in dynamic, non-linear biological processes without appropriate adaptation. Immunologists must gain foundational AI knowledge to effectively evaluate computational models and select appropriate techniques for their research.

AI modeling in immunology frequently follows a reductionist approach, wherein specific biological processes are extracted and formalized into computational algorithms. While this allows for detailed analysis of immune interactions, it can sometimes result in models that fail to capture the complexity of immune dynamics. Despite AI's ability to analyze immune system behaviors and predict outcomes, translating these insights into real immunological applications remains a challenge. Many AI-generated predictions require further refinement before they can be applied in clinical or experimental settings. Some models suffer from excessive parameterization, where unnecessary variables obscure biological relevance, leading to misinterpretations of experimental data. Additionally, the effectiveness of AI models is often limited by the quality and completeness of the input data. Poorly annotated datasets or insufficient immunological information can lead to inaccurate computational predictions. To improve AI integration in immunology, enhanced communication between AI researchers and immunologists is crucial. AI researchers must engage with experimental immunology literature, understand the limitations of immunological assays, and design models that align with real-world biological mechanisms. AI models often prioritize efficiency by minimizing the number of variables considered, frequently reducing immune system simulations to two-dimensional dynamic models (e.g., protein concentration over time). Few AI models incorporate spatial dimensions of immune responses, as spatial modeling significantly increases computational complexity. Furthermore, immunological spatial datasets remain limited, making AI-driven spatial modeling more challenging. Immunologists must emphasize that both the position and movement of immune-related molecules are critical for any AI-driven model. While AI-driven immunology often focuses on cellular interactions, future models should integrate immune tissue environments and lymphoid/non-lymphoid organ dynamics. AI-driven models that consider only immune cell counts may overlook crucial spatial interactions that impact immune responses. The lack of understanding among AI researchers regarding immune cell dynamics, protein transport pathways, and tissue-specific interactions remains a significant challenge.

AI-based models can incorporate multivariate immune cell populations; however, increasing the complexity of the model may reduce interpretability and usability. AI researchers often favor simpler models that yield interpretable outputs, while immunologists require models that accurately reflect real immune system behaviors. This trade-off must be carefully managed to ensure AI models remain useful and clinically relevant. Many AI-driven models make naïve assumptions that oversimplify the complexity of immune responses. For example, some AI models treat T cell immunity as consisting solely of effector $CD4^+$ or $CD8^+$ T cells, failing to account for regulatory T cells ($CD4^+CD25^+$), which play a crucial role in balancing immune activation and suppression. While similar AI modeling approaches have been successfully applied in fields like chemical engineering (e.g., modeling population dynamics in industrial bioreactors), their application to immunology requires additional complexity. The primary goal of AI modeling in immunology is to generate hypotheses and identify key experimental variables. AI models should guide experimental immunologists toward promising research directions by highlighting immune mechanisms that warrant further investigation. Testing and refining AI-driven models with experimental data will improve their reliability and applicability in immunology research. AI has been increasingly applied in studying the interactions between HIV, HSV, and immune target cells. While AI-driven models have provided valuable insights into AIDS and herpes disease progression, most models assume that target cells are infected with a single virus (e.g., either HIV or HSV). This assumption fails to reflect real-world co-infection scenarios, where cells can be simultaneously infected by multiple viruses. AI-driven models must evolve to better simulate multi-pathogen interactions and the immune system's response to complex infections. To account for co-infections, AI modeling must integrate data on how different pathogens interact within the immune environment. Traditional AI models struggle to capture the full dynamics of viral replication, immune evasion strategies, and immunotherapeutic interventions. Future models must incorporate real-world biological complexities to accurately simulate disease progression and treatment outcomes.

AI researchers serve as translators of experimental immunology findings, converting empirical data into computational frameworks. However, for AI models to be meaningful, AI researchers must develop a solid foundation in modern immunological principles. This will allow them to formulate biologically relevant hypotheses and develop models that reflect true immune system behavior. Within AI research, there are distinct groups: (a) theoretical AI researchers who focus on algorithm development with little interest in immunology, (b) applied AI researchers who aim to integrate AI into biomedical sciences but often lack deep immunology knowledge, and (c) computational biologists who specialize in translating immunological data into AI models. The latter group is best suited to drive AI-driven immunological research forward. Regardless of expertise, AI researchers must dedicate significant time to understanding immunological mechanisms to create meaningful models. The complexity of the immune system presents challenges not only for AI researchers but also for immunologists themselves. Even with advances in experimental techniques, immunologists frequently encounter unexpected immune behaviors that require continuous reevaluation of existing theories. AI can aid in identifying patterns and refining immunological hypotheses, but collaboration with experimental immunologists is essential to validate AI-generated insights. Unlike traditional fields such as physics and engineering, AI has only recently been applied to immunology. Despite working in related disciplines, AI researchers and immunologists have historically operated independently. To bridge this gap, interdisciplinary teams should consist of immunologists, computational biologists, and AI researchers who collaboratively develop AI-driven immunological models.

Ultimately, AI-driven immunology holds immense potential, but its success depends on interdisciplinary collaboration, careful validation, and continuous refinement of computational models. AI should be viewed as a tool to augment, not replace, traditional immunological research. By integrating AI into immunology, researchers can accelerate discoveries, enhance our understanding of immune responses, and improve immunotherapeutic strategies and patient outcomes.

**Support and Conflict of Interest:** LBM has an equity interest in TechImmune, LLC., a company that may potentially benefit from the research results and serves on the company's Scientific Advisory Board. LBM's relationship with TechImmune, LLC., has been reviewed and approved by the University of California, Irvine under its conflict-of-interest policies.

**Acknowledgements**: Studies of this report were supported by Public Health Service Research grants AI158060, AI150091, AI143348, AI147499, AI143326, AI138764, AI124911, and AI110902 from the National Institutes of Allergy and Infectious Diseases (NIAID) and grants EY19896, EY14900, EY14017, EY09392  from the National Eye Institutes to LBM.

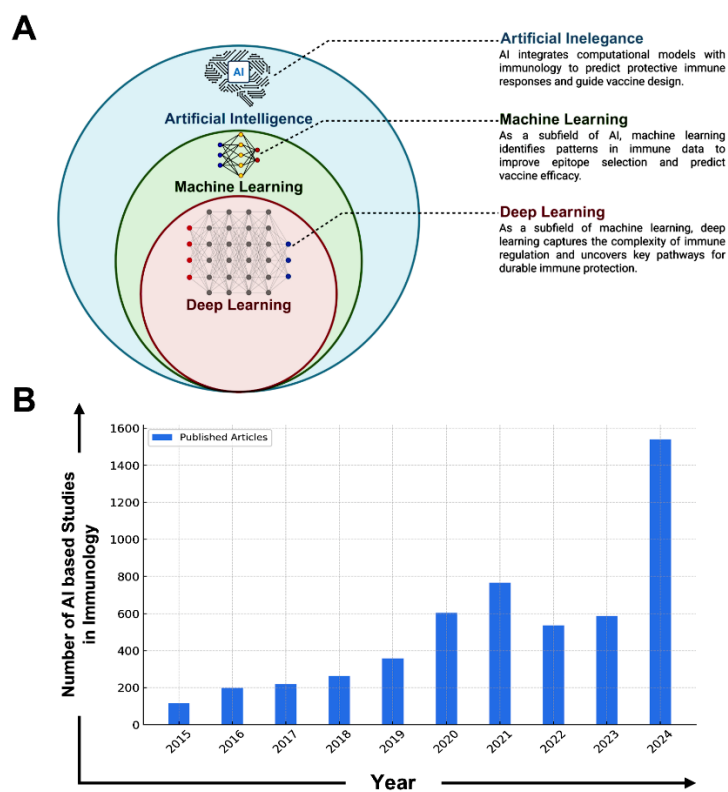

**Figure 1.** **Growth of AI-Driven Research in Immunology (2015–2024). (A)** Conceptual representation of the hierarchical relationship among Artificial Intelligence (AI), Machine Learning (ML), and Deep Learning (DL) in the context of immunological applications. AI serves as the broadest category, encompassing ML techniques that identify patterns in immune-related datasets. DL, a specialized subfield of ML, is used to model complex immune responses and uncover pathways critical for vaccine efficacy and immune regulation. (**B**) The integration of Artificial Intelligence (AI) in immunology has expanded significantly over the past decade. The number of publications in this field has grown from just over 100 articles per year in 2015 to over 1500 in 2024, highlighting the increasing role of AI in immunological research and its applications in vaccine design, immune response modeling, and precision medicine. This diagram was generated based on keyword searches such as "AI in immunology," "machine learning in immunology," and "deep learning in immune research" from the PubMed database. The search encompassed peer-reviewed articles, systematic reviews, and conference proceedings related to AI-driven immunological studies. The data reflects the rising trend in AI applications, including areas such as predictive immune modeling, AI-assisted diagnostics, and computational vaccine development. The exponential rise in publications post-2020 aligns with breakthroughs in deep learning, large-scale immunological datasets, and AI-driven drug discovery, indicating a paradigm shift in how computational tools are being utilized in immunology. The projected increase in AI-based studies suggests continued advancements in immune system modeling, personalized immunotherapy, and AI-enhanced vaccine development strategies.

El Fatimi *et al.* Figure 2

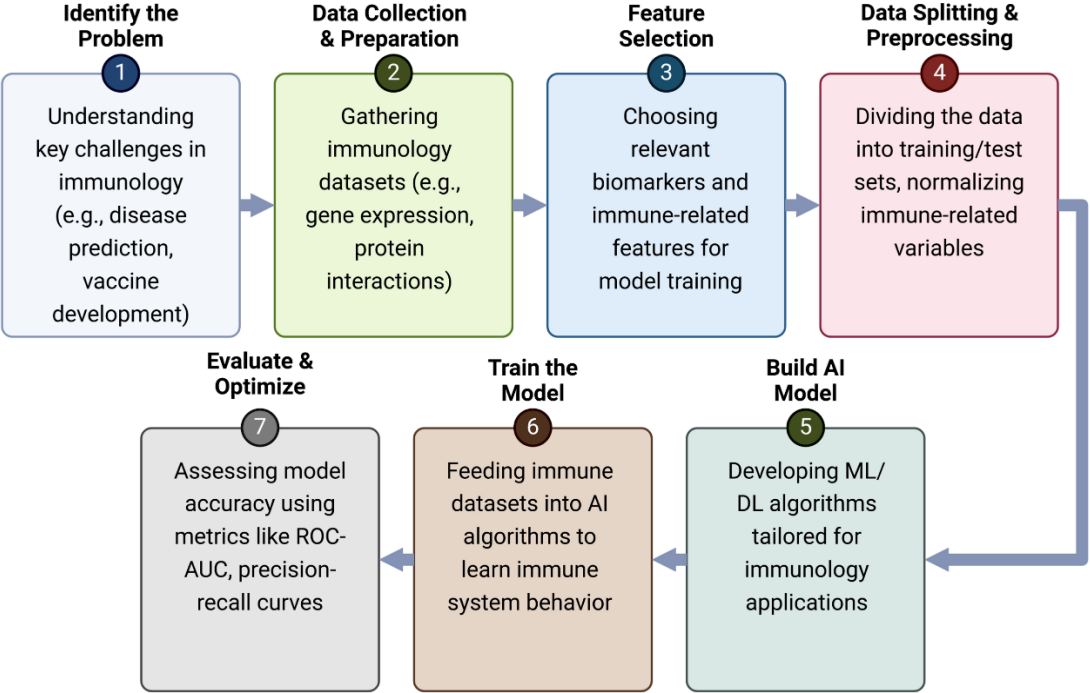

**Figure 2**. **AI-Driven workflow for immunology research**. This illustration presents a structured

framework outlining the role of Artificial Intelligence (AI) in immunology research, demonstrating how AI-driven approaches enhance immunological analysis, predictive modeling, and therapeutic development. The process begins with the identification of key immunological challenges, such as disease prediction and vaccine development, followed by the collection and preprocessing of immunological datasets, including gene expression profiles and protein interactions. Feature selection then plays a crucial role in identifying relevant biomarkers and immune-related variables to improve model accuracy. Once the data is prepared, machine learning and deep learning models are built and trained on immune system datasets to recognize patterns and predict immune responses. The trained models undergo evaluation and optimization using performance metrics like ROC-AUC and precision-recall curves to ensure accuracy and reliability. Ultimately, these AI-driven models are applied in real-world immunology research, enabling advancements in personalized medicine, immunotherapy, and vaccine optimization. By integrating AI techniques into immunology, researchers can gain deeper insights into immune system behavior, develop more effective therapeutic interventions, and refine disease treatment strategies.

El Fatimi *et al.* Figure 3

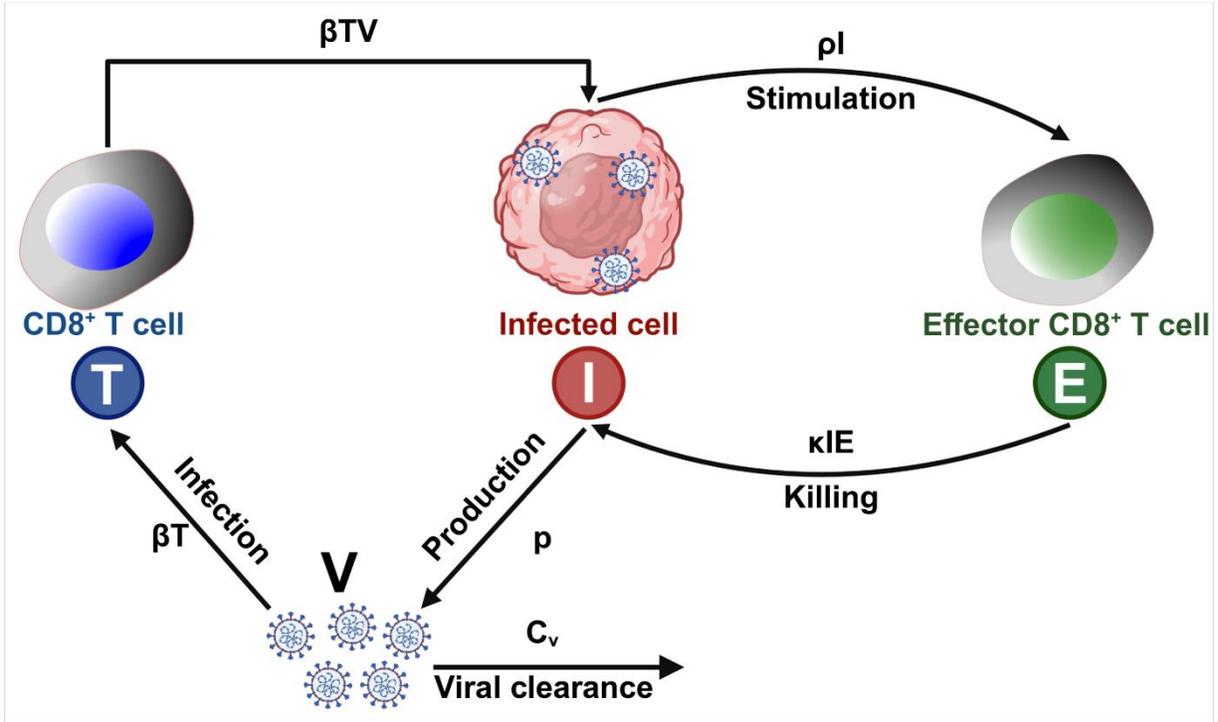

**Figure 3**. **AI-Driven Mathematical Model of CD8+ T Cell Response to Viral Infections**: **T (Blue)** CD8[+] T cells that recognize and eliminate infected cells**; I (Red)** – Virus-infected cells that replicate and produce viral particles; **E (Green)** – Effector immune cells (e.g., activated CD8+ T cells); **V** – Free virus particles that infect host cells; βT – Rate at which virus infects host cells and generates infected cells (I); βTV – Interaction coefficient representing how CD8+ T cells (T) are stimulated by viral load (V); **p** – Rate of virus production by infected cells (I); **κIE** – Killing rate of infected cells (I) by effector T cells (E); **ρI** – Stimulation coefficient, representing the activation of effector immune cells (E) by infected cells (I); **Cv** – Clearance rate of virus (V) due to immune response.

El Fatimi *et al.* Figure 4

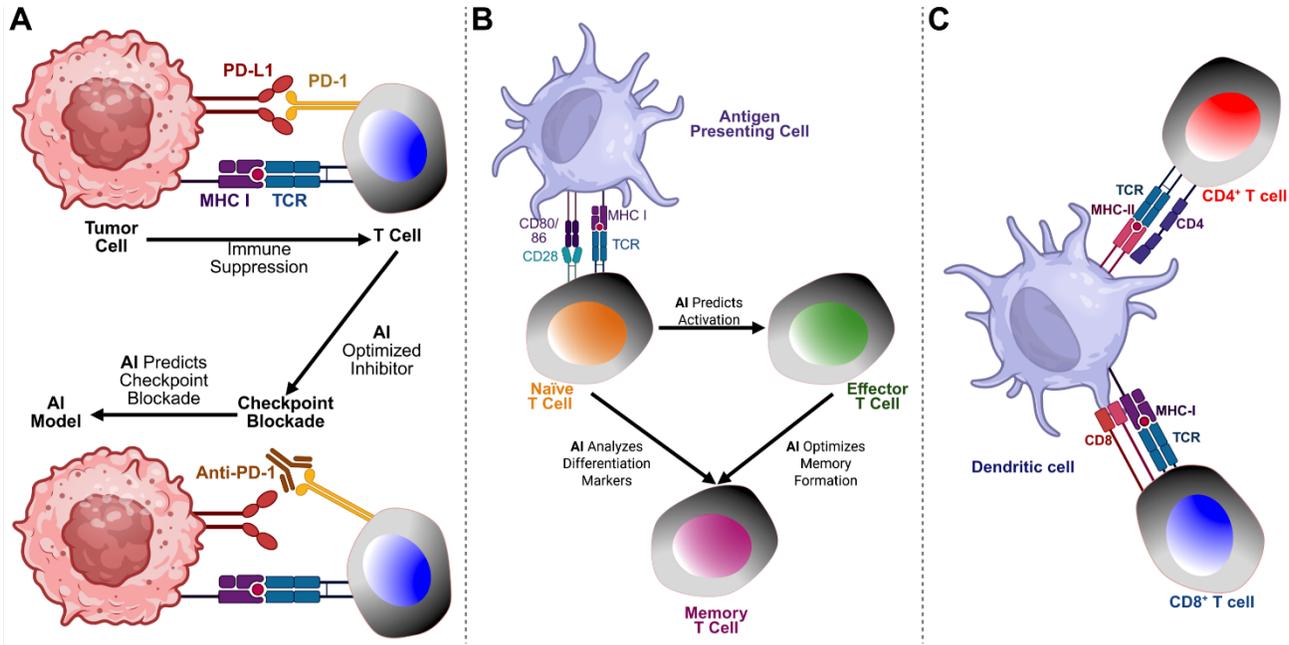

**Figure 4.** AI-Driven Insights into Tumor-Immune Interactions and Memory T Cell Differentiation. **(A)** AI's Role in Tumor-Immune Interactions and Checkpoint Blockade: This panel illustrates how Artificial Intelligence (AI) supports understanding and optimization of tumor-immune interactions, particularly in regulating immune checkpoint inhibitors. Tumor cells suppress T cell activity through checkpoint pathways such as PD-1/PD-L1 and CTLA-4. AI models predict optimal checkpoint blockade strategies by identifying patient-specific responses, discovering biomarkers, and enhancing combination immunotherapies. Integration of multi-omics data (genomics, transcriptomics, proteomics) via AI enables personalized immunotherapy with improved efficacy and reduced toxicity. **(B)** AI Insights into Memory T Cell Differentiation: This panel demonstrates AI's role in guiding memory T cell differentiation. Following antigen exposure, naïve T cells activate and differentiate into effector and memory T cells. AI enhances this process by predicting activation based on TCR signaling, analyzing differentiation markers, and optimizing memory formation. AI models trained on single-cell RNA sequencing and epigenetic data uncover pathways critical for long-term immune protection, informing vaccine development and cancer immunotherapies. **(C)** Interaction Between Dendritic Cells and T Cells in Antigen Presentation. This illustration depicts the crucial role of dendritic cells (DCs) in bridging innate and adaptive immunity through antigen presentation. The dendritic cell presents antigens to CD4[+] T cells via MHC class II molecules, recognized by the T cell receptor (TCR) in conjunction with the CD4 co-receptor, facilitating helper T cell activation. Simultaneously, CD8[+] T cells recognize antigens presented on MHC class I molecules, with the TCR engaging the complex alongside the CD8 co-receptor, leading to cytotoxic T cell activation. This dual interaction is essential for initiating and coordinating immune responses, enabling helper T cells to support other immune cells and cytotoxic T cells to eliminate infected or malignant cells.

El Fatimi *et al.* Figure 5

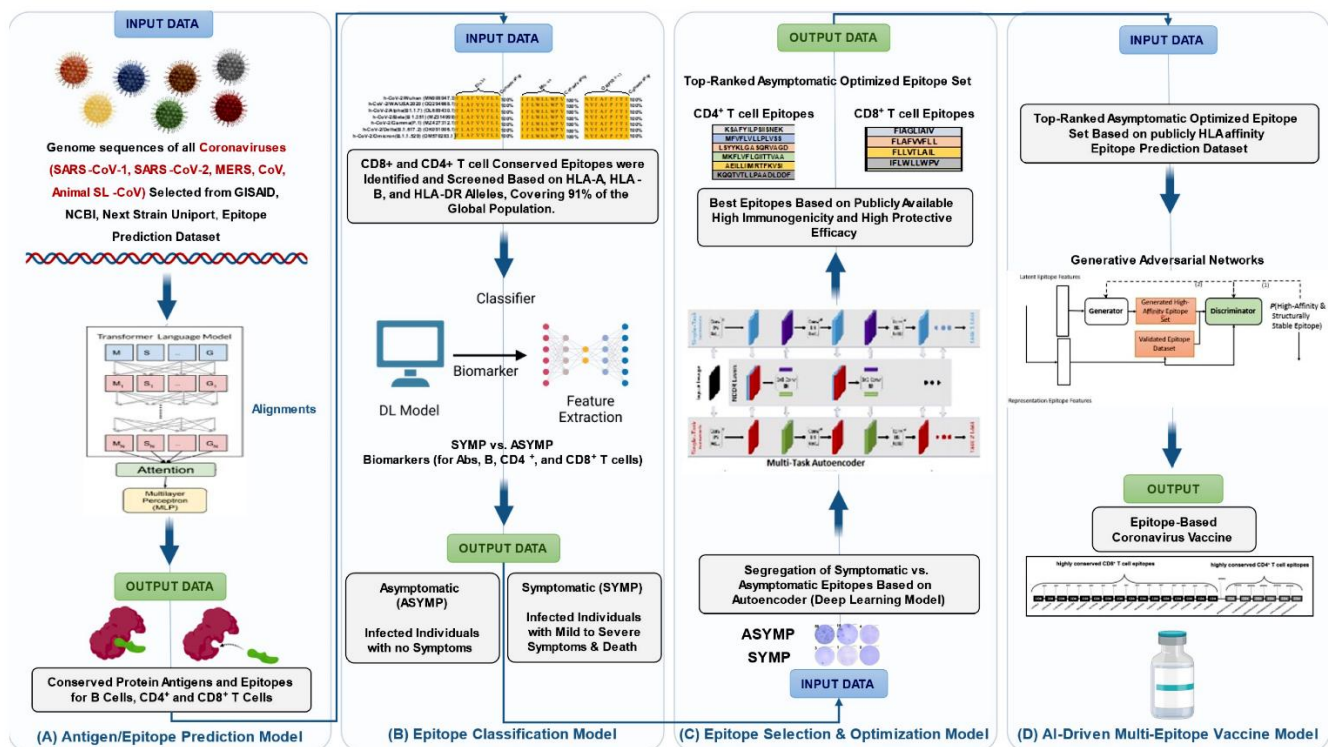

**Figure 5.** AI-driven framework optimizes B- and T-cell epitope prediction, classification, and multi-epitope vaccine design. (**A**) Epitope Prediction Model (Transformer-based) for identifying immunogenic B-cell and T-cell epitopes, (**B**) Deep learning epitope classification using (CNN-based) biomarker extraction and feature selection, (**C**) A multi-task autoencoder prioritizes immunogenic and protective epitopes, and (**D**) AI-powered vaccine formulation integrating top-ranked asymptomatic B- and T-cell epitopes into a candidate vaccine using Generative Adversarial Networks (GANs). GANs refine and generate four multi-epitope-based next-generation coronavirus vaccine candidates.

El Fatimi *et al.* Figure 6

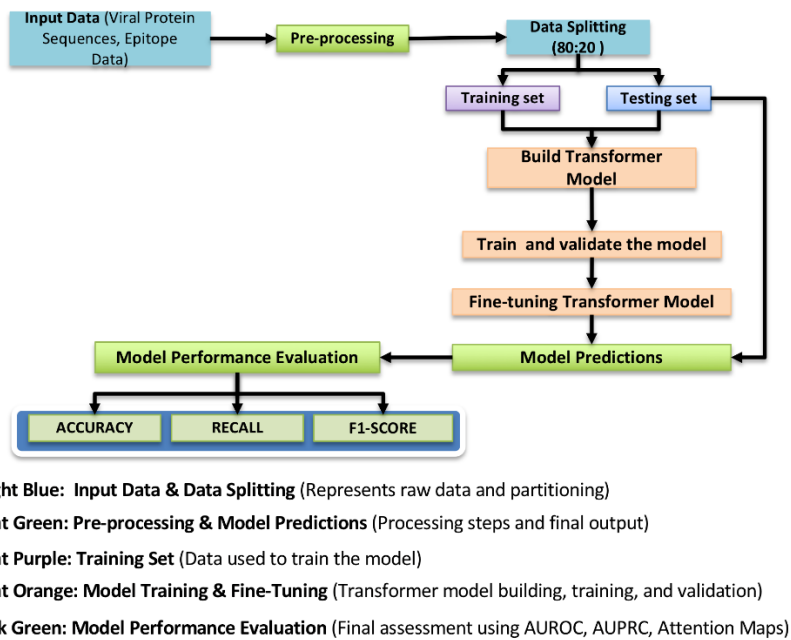

🔵 **Light Blue: Input Data & Data Splitting** (Represents raw data and partitioning)
🟩 **Light Green: Pre-processing & Model Predictions** (Processing steps and final output)
🟪 **Light Purple: Training Set** (Data used to train the model)
🟧 **Light Orange: Model Training & Fine-Tuning** (Transformer model building, training, and validation)
🟩 **Dark Green: Model Performance Evaluation** (Final assessment using AUROC, AUPRC, Attention Maps)

**<u>Figure 6</u>. Antigen/Epitope Prediction Model Workflow**. This figure illustrates the structured workflow of the transformer-based deep learning model used for antigen/epitope prediction. The pipeline begins with input data, consisting of viral protein sequences and epitope data, which undergo preprocessing to extract relevant features. The dataset is then split into training (80%) and testing (20%) subsets to ensure proper model generalization. The transformer-based model is trained and fine-tuned using deep learning techniques to improve prediction accuracy. Finally, the model undergoes performance evaluation using metrics such as Accuracy, Recall, and F1-score, providing insight into the reliability of predicted immunogenicity scores. This structured approach ensures that the model accurately differentiates highly immunogenic and non-immunogenic epitopes, aiding in vaccine target selection.

El Fatimi *et al.* Figure 7

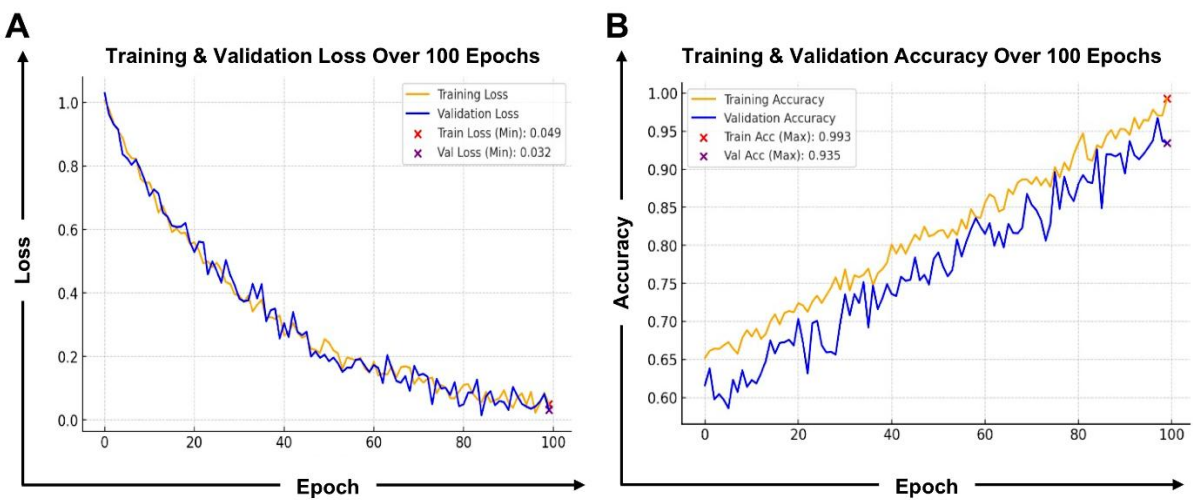

**Figure 7.** **Training and Validation Performance of the Antigen/Epitope Prediction Model: (A) Loss and (B) Accuracy.** This figure presents the training performance metrics of the antigen/epitope prediction model. The left graph displays the training and validation loss over 100 epochs, where both losses decrease progressively, indicating stable optimization. The validation loss stabilizes at 0.032, demonstrating strong generalization without overfitting. The right graph showcases training and validation accuracy, which steadily improve, reaching 0.993 training accuracy and 0.935 validation accuracy. These results confirm that the model effectively learns to recognize key features of immunogenic and non-immunogenic epitopes while maintaining high reliability across validation data. The minimal gap between training and validation curves highlights robust model performance, ensuring applicability for real-world antigen screening.

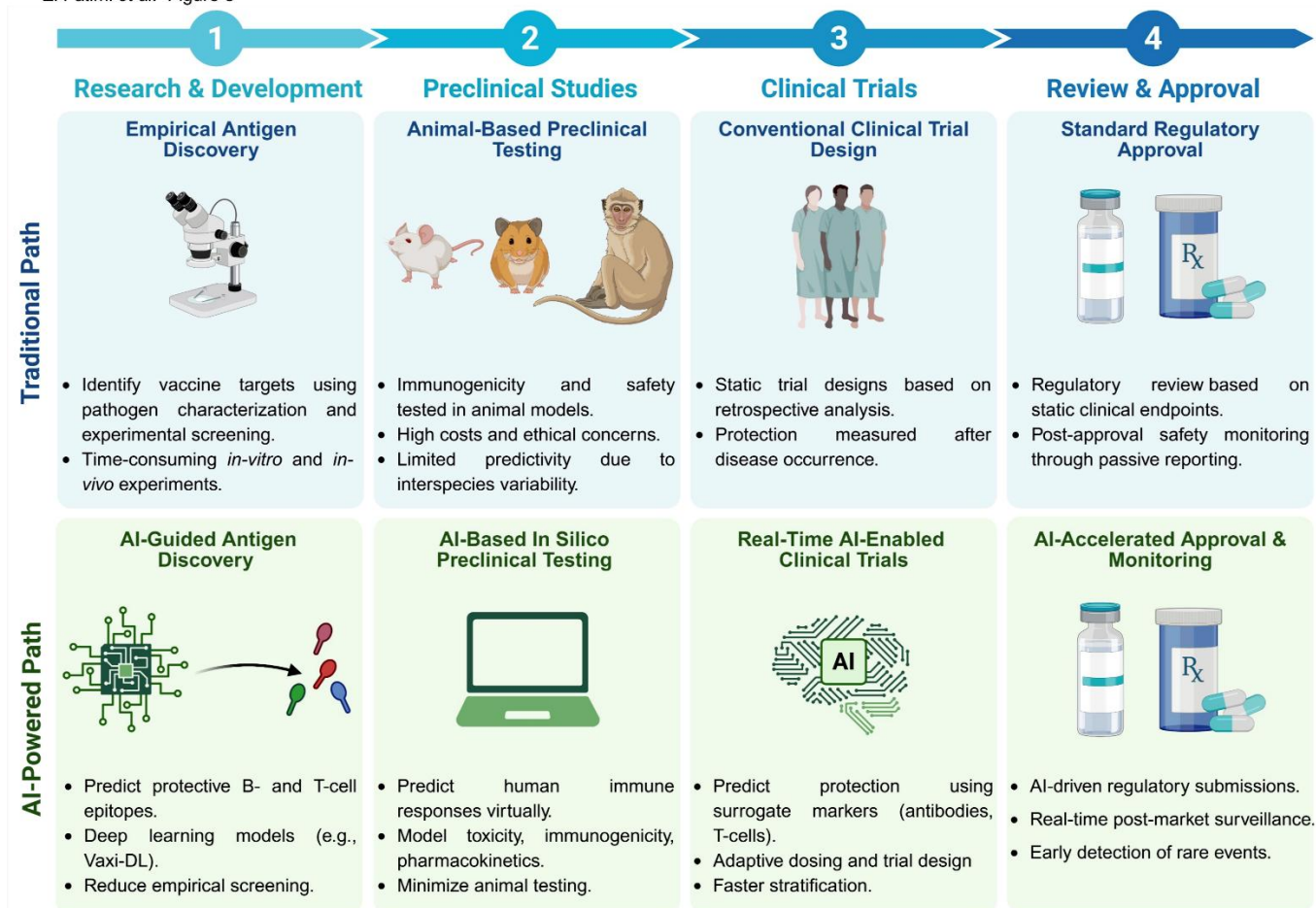

**Figure 8.** **Artificial intelligence revolutionizing preclinical and clinical development of vaccines and immunotherapeutics.** The traditional pathway (top panels) relies heavily on empirical antigen discovery, animal-based preclinical testing, conventional clinical trial designs, and retrospective regulatory evaluation, leading to ethical concerns, interspecies variability, and prolonged development timelines. In contrast, the AI-powered pathway (bottom panels) illustrates how emerging computational models can replace or reduce the need for animal preclinical testing, as endorsed by the United States FDA under the Modernization Act 2.0. AI-driven platforms leverage deep learning to predict protective antigens, simulate human immune responses *in-silico*, and enable real-time immunobridging strategies during clinical trials. These advancements accelerate candidate selection, optimize clinical trial design, predict protection based

on immune markers, and enhance the speed and precision of regulatory approval and post-market surveillance.

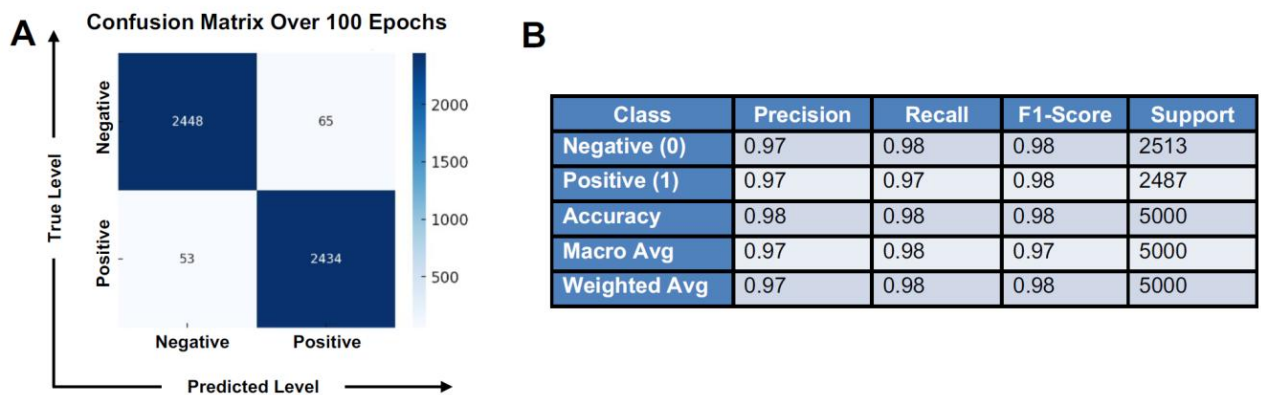

**Figure S1. (A)** Confusion Matrix for Antigen Prediction Model Over 100 Epochs. This confusion matrix provides an in-depth evaluation of the classification performance of the antigen/epitope prediction model. The matrix indicates that the model correctly classified 2448 negative samples (non-immunogenic epitopes) and 2434 positive samples (immunogenic epitopes). There were only 65 false negatives (immunogenic epitopes misclassified as non-immunogenic) and 53 false positives (non-immunogenic epitopes incorrectly predicted as immunogenic). These results highlight the model's high precision and recall, reducing the likelihood of false predictions. A low false-negative rate ensures that strong epitopes are not mistakenly omitted from vaccine candidate selection, while a low false-positive rate prevents the misclassification of weak epitopes, ensuring accurate antigen screening for immunotherapy and vaccine research. and (B) Classification Report for Antigen Prediction Model Over 100 Epochs.

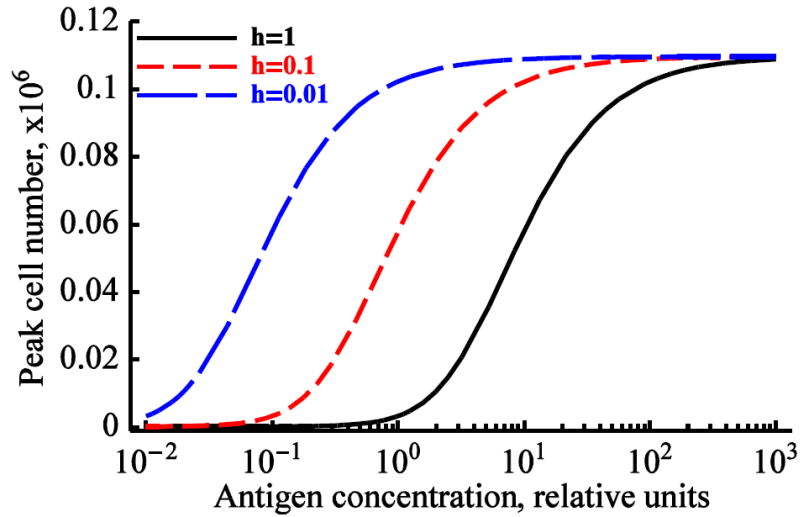

**Figure S3. Peak Immune Response as a Function of Antigen Concentration.** The immune response is highly dependent on antigen concentration, as illustrated in Figure 2, which depicts the prediction of T cell proliferation based on antigen availability. The model assumes that T cell expansion follows a saturating function, where the rate of proliferation (ρ) is governed by antigen concentration (I) and a half-saturation constant (h). At low antigen concentrations, T cell activation remains minimal, but as antigen levels increase, proliferation accelerates, reaching a peak when antigen availability is optimal. However, beyond a certain threshold, further increases in antigen concentration do not significantly enhance proliferation, reflecting a saturation effect in immune activation. This model assumes an initial population with a proliferation rate (ρ) of 1 division per day, sustained over one week. Such models are critical for optimizing vaccine formulations and immunotherapy strategies, as they help predict optimal antigen dosing to maximize immune activation while minimizing the risks of T cell exhaustion. AI-driven modeling approaches can further refine these predictions by incorporating real-time patient data, enhancing precision in immune response forecasting and personalized treatment strategies.